\documentclass[runningheads,a4paper]{llncs}
\usepackage{amsmath}
\usepackage{amssymb}
\usepackage{graphicx}
\usepackage{bm}
\usepackage{verbatim}

\usepackage{url}

\newcommand{\keywords}[1]{\par\addvspace\baselineskip
\noindent\keywordname\enspace\ignorespaces#1}

\def\bi{\begin{itemize}}
\def\ei{\end{itemize}}
\def\be{\begin{equation}}
\def\ee{\end{equation}}

\def\b{{\mathbf b}}

\def\v{{\mathbf v}}
\def\x{{\mathbf x}}
\def\m{{\mathbf m}}

\def\h{{\mathbf h}}
\def\m{{\mathbf m}}

\newcommand{\diffd}{\mathrm{d}}

\def\VB{\bm{v}^{\mathrm{B}}}
\def\VF{\bm{v}^{\mathrm{F}}}
\def\vB{v^{\mathrm{B}}}
\def\vF{v^{\mathrm{F}}}

\def\HB{\bm{h}^{\mathrm{B}}}
\def\HF{\bm{h}^{\mathrm{F}}}
\def\hB{h^{\mathrm{B}}}
\def\hF{h^{\mathrm{F}}}
\def\PFG{p_{\mathrm{FG}}}
\def\PBG{p_{\mathrm{BG}}}

\def\ns{\negthickspace}
\begin{document}

\mainmatter  % start of an individual contribution

% first the title is needed
\title{Weakly Supervised Learning of Foreground-Background Segmentation using Masked RBMs}

% a short form should be given in case it is too long for the running head
\titlerunning{Masked RBMs for Foreground-Background Segmentation}

% the name(s) of the author(s) follow(s) next
%
% NB: Chinese authors should write their first names(s) in front of
% their surnames. This ensures that the names appear correctly in
% the running heads and the author index.
%
\author{Nicolas Heess\inst{1} \and Nicolas Le Roux\inst{2} \and John Winn\inst{3}}
\authorrunning{Masked-RBMs for Foreground-Background Segmentation}
% (feature abused for this document to repeat the title also on left hand pages)

% the affiliations are given next; don't give your e-mail address
% unless you accept that it will be published
\institute{University of Edinburgh, IANC, Edinburgh, UK\\
\and
INRIA, Sierra Team, Paris, France
\and
Microsoft Research, Cambridge, UK}

%
% NB: a more complex sample for affiliations and the mapping to the
% corresponding authors can be found in the file "llncs.dem"
% (search for the string "\mainmatter" where a contribution starts).
% "llncs.dem" accompanies the document class "llncs.cls".
%

\maketitle

\begin{abstract}
We propose an extension of the Restricted Boltzmann Machine
(RBM) that allows the joint shape and appearance of foreground objects
in cluttered images to be modeled independently of the background. We
present a learning scheme that learns this representation directly
from cluttered images with only very weak supervision. The
model generates plausible samples and performs foreground-background
segmentation. We demonstrate that representing foreground objects
independently of the background can be beneficial in recognition tasks.

\end{abstract}

\section{Introduction}
Learning generative models of natural images is a long-standing
challenge. Recently, a new spectrum of approaches, loosely referred to
as ``deep learning'' (DL), has led to advances in several AI-style learning
tasks. At the heart of this framework is the use of RBMs for greedy
learning of multi-layered representations
such as Deep Belief Networks (DBN,~\cite{HintonNC2006}) and Deep
Boltzmann Machines (DBM,~\cite{Salakhutdinov2009}). However, despite
interesting applications in vision (e.g.~\cite{LeeICML2009}),
it has become apparent that the basic formulation of the RBM is too
limited to model images well, and several alternative formulations have
recently been proposed (e.g.~\cite{RanzatoCVPR2010}).
One powerful notion from the computer vision literature is that of a
layered representation. This allows images to be composed
from several independent objects and can account for occlusion
(e.g.~\cite{WangTIP1994,WilliamsNC2004}). In~\cite{LeRouxMSR2010}, we
have proposed a model that introduces such a
layered representation into the DL framework. In this
model, the Masked RBM (MRBM), an image is
composed of several regions, each of which is modeled in terms of its
shape and appearance. The region shape determines where a region is
visible, and the appearance determines the color or
texture of the region, while overlapping regions occlude each
other. We used this architecture to formulate a generative model of
lower-level structure
in \emph{generic} images
and therefore assumed that all regions were equivalent, i.e. all
regions were governed by the same shape and appearance models, and that
shape and appearance were independent. For higher-level
tasks such as recognition, however, the different regions of an image
are not equivalent and some are more interesting than others.
Here, we show that separating an image into
\textit{qualitatively different} layers and \textit{jointly}
modeling shape and appearance allows us to learn and represent
specific objects or object categories \textit{independently} of the
background. As a result, the representation of the foreground
is less affected by background clutter. Our model is able to perform
foreground-background segmentation, and it can generate new
instances of the foreground objects (shape and appearance).
In particular, we show that learning is possible directly from cluttered
images and requires only very weak supervision: inspired by
\cite{WilliamsNC2004}, we bootstrap learning with an approximate model
of the background. This is easily obtained by training on general
natural images and sufficient for learning then to proceed without further
supervision: foreground objects can be detected as outliers under the
background model and a model of the foreground can thus be learned
from the regularities of these outliers across training images. To our
knowledge, foreground-background segmentation has not
previously been addressed in the DL framework.
% primarily because standard RBMs (and DBNs/DBMs) are not directly
% suitable for this purpose.
Tang \cite{TangNIPS2010} proposes a model that is related to ours
and applies it to the problem of recognition under
occlusion, but considers a simpler scenario with binary images
and fully supervised learning only.

%%%%%%%%%%%%%%%%%%%%%%%%%%%%%%%%%%%%%%%%%%%%%%%%%%%%%%%%%%%%%%%%%%%%%%
%%%%%%%%%%%%%%%%%%%%%%%%%%%%%%%%%%%%%%%%%%%%%%%%%%%%%%%%%%%%%%%%%%%%%%
\section{Model}
\label{sec:Model}
Our model extends the MRBM presented in~\cite{LeRouxMSR2010}: instead
of modeling general images that consist of generic and equivalent
regions, it assumes that images
contain a single foreground object in front of a cluttered background
(which can and often will also contain parts of other
objects, but these are not explicitly modeled). Foreground and
background are assumed to be independent and the background is
occluded by the foreground object. In the model, this is achieved by
composing the \textit{observed} image from two \textit{latent} images: one
for the background, and one for the foreground. The background image
is visible only where the foreground is not, and the visibility of the
foreground image is determined by a binary mask. Intuitively speaking,
the foreground image determines the \textit{appearance} of the
foreground object, and the mask determines its \textit{shape}.
The model is a pixel-wise binary mixture with the mixture
component for each pixel specified by the mask. Denoting the
observed image by $\x$, the background image by $\VB$, and the
appearance and shape of the foreground by $\VF$ and $\m$
respectively, the model can be written as
\begin{equation}
P(\x) \ns = \ns \sum_{\m} \ns \int \ns \ns \diffd \VB \ns \ns \int \ns \diffd \VF
\ns \left(\!\prod_i \ns\delta \!\left[\vF_i \ns = \ns x_i\right]^{m_i}
\ns \!\delta\!\left[\vB_i \ns = \ns x_i\right]^{(1-m_i)}\ns \right)
\ns p_{\mathrm{FG}}(\VF\ns, \m) p_{\mathrm{BG}}(\VB)
\label{eq:FullModel}
\end{equation}
where $i$ is the pixel index, $m_i \in \{0,1\}$,
and the product of delta functions forces, for each pixel, one of the
two latent images to take on the value of the observed image at that
pixel. The mask determines which latent image is chosen.
We formulate the priors over the background
image and over foreground appearance and shape as RBMs.
Assuming that pixels are continuous valued in $[0,1]$, we
choose for $p_{\mathrm{BG}}$ a special form of the Beta RBM
with energy $E_{\mathrm{Beta}}(\VB, \h)$ described in
\cite{LeRouxMSR2010}
so that $p_{\mathrm{BG}}(\VB) = 1/Z \sum_{\h}
\exp\{ -E_{\mathrm{Beta}}( \VB, \h; \Theta_{\mathrm{BG}})\}$.
Unlike the Gaussian RBM with fixed variance commonly used for continuous data
(e.g.~\cite{LeeICML2009}), the Beta RBM models mean \emph{and}
variance of the visible units.

The model of the foreground object defines a joint distribution over a
continuous valued image and a binary mask.
Here, we are interested in the case where shape and appearance are \textit{dependent} and we will therefore model them jointly.
This is in contrast with the approach taken in~\cite{LeRouxMSR2010} where they were assumed to be independent,
i.e. $p_{\mathrm{FG}}(\VF,\m) =
p_{\mathrm{FG}}(\VF)p_{\mathrm{FG}}(\m)$.
%Since the shape and
%appearance of the foreground are heavily \textit{dependent}, we will
%model them jointly. This is in contrast with the approach taken
%in~\cite{LeRouxMSR2010}, where they were independent,
%i.e. $p_{\mathrm{FG}}(\VF,\m) =
%p_{\mathrm{FG}}(\VF)p_{\mathrm{FG}}(\m)$.
Thus, in this new model, we use a particular form of the RBM which has two sets of visible units, a set of binary units for the mask $\m$ and a set of continuous
valued units for the appearance image $\VF$:
\begin{equation}
E_{\mathrm{mixed}}(\v,\m,\h;\Theta) =
E_{\mathrm{Bin}}(\m,\h;\Theta^{\mathrm{S}}) +
E_{\mathrm{Beta}}(\v,\h;\Theta^{\mathrm{A}}) \quad ,
\end{equation}
where $E_{\mathrm{Bin}}(\m,\h;\Theta)\ns =\ns \m^TW\h + \b^T \! \m$ is the
energy function of a binary RBM, and the joint distribution is thus
given by: $p_{\mathrm{FG}}(\VF \ns ,\m)\ns =\ns \frac{1}{Z} \sum_{\h} \exp\{
-E_{\mathrm{mixed}}(\VF\ns,\m,\!\h;\Theta) \}$.

\textbf{Inference}: although exact inference is intractable, an efficient
Gibbs sampling scheme exists, as detailed in~\cite{LeRouxMSR2010}.
Let
$\HF$ and $\HB$ denote the hidden units of the foreground and the
background model respectively, then the following properties admit a
Gibbs sampling scheme in which the three sets of variables
$(\HF,\HB)$, $\m$, and $(\VF, \VB)$ are sampled in turn:
\begin{enumerate}
\item given $\VF$ and $\VB$ the hidden units $\HF$ and $\HB$ are
conditionally independent, i.e. $p(\HF|\VF) = \prod_j p(\hF_j | \VF)$ and
$p(\HB | \VB) = \prod_j p(\hB_j | \VB)$~,
\item given the hidden variables $\HF$, $\HB$ and the image $\x$, the
variables of the mask, foreground image, and background image are
pixel-wise conditionally independent, i.e.
$p(\VF,\VB,\m | \HF,\HB, \x) = \prod_i p(\vF_i, \vB_i, m_i | \HF, \HB,
\x)$~,
\item $p(\vF_i, \vB_i, m_i | \HF, \HB, \x)$ can be decomposed as
follows
\begin{align}
p(\vF_i \!,\vB_i\!, m_i | \HF \ns, \HB \ns, \x)\! &=\! p(\vF_i, \vB_i |
m_i, \HF, \HB, \x)p(m_i | \HF\!, \HB \!, \x) \label{eq:conditional1}
\\
p(m_i = 1| \HF\ns, \HB\ns, \x)\! &=\! \frac{\PFG(\vF_i\ns =\! x_i, m_i\ns =\! 1 | \HF)}
{\PBG(\vB_i\ns =\! x_i | \HB) \PFG( m_i\ns =\! 0 | \HF\!)\ns + \ns
\PFG(\vF_i\ns =\! x_i, \! m_i\ns =\! 1 |\HF\!) } \label{eq:conditional2} \\
p(\vF_i\!, \vB_i | m_i, \HF \ns, \HB \ns, \x)\! &=\! \left\{
\begin{array}{ll}
\delta\left[\vF_i = x_i\right] \PBG(\vB_i|\HB) & ~~\mathrm{if}~ m_i = 1 \\
\delta\left[\vB_i = x_i\right] \PFG(\vF_i|\HF) & ~~\mathrm{otherwise.}
\end{array} \label{eq:conditional3} \right.
\end{align}
\end{enumerate}

\textbf{Learning}: during learning with unlabeled data only $\x$ is observed. We use an
EM-like approach in which inference of $\VF$, $\VB$, and $\m$
alternates with updates of the model parameters. Once these
variables have been inferred, they can be treated as ``observed''
data for one of the usual learning schemes for RBMs such as
contrastive divergence (CD,~\cite{HintonNC2002}) or stochastic maximum
likelihood
(SML, also referred to as ``persistent CD''~\cite{TielemanICML2008}),
which we use in the experiments below.
SML relies on persistent chains of samples to represent the
model distribution which are updated by one step of Gibbs sampling
per iteration. Note that, due to the directed nature of the
mixture in (\ref{eq:FullModel}), the persistent Markov chains
representing the model distributions of the
two RBMs for foreground and background do not interact, i.e. we run
independent chains as if we were training both RBMs separately.
Fully unsupervised learning is possible in principle but likely
to be very difficult for all but very simple datasets. We therefore
consider a ``weakly supervised'' scenario related to
\cite{WilliamsNC2004}: we assume that we
have some general knowledge of the statistical regularities to be
expected in the \emph{background}. This approximate model of the
background can be obtained by training on general natural images
and it allows us to bootstrap learning of the foreground
model from unsegmented training data. Foreground objects stand
out from the background, i.e. they appear as ``outliers'' to the
background model which forces them to be explained by the foreground
model. Although this detection is initially unreliable, the foreground
model can then learn about the consistencies of
these outliers across the training images which eventually leads
to a good model of the foreground objects without any additional
information being provided or prior knowledge (e.g. coherence or
convexity) being used.

%%%%%%%%%%%%%%%%%%%%%%%%%%%%%%%%%%%%%%%%%%%%%%%%%%%%%%%%%%%%%%%%%%%%%%
%%%%%%%%%%%%%%%%%%%%%%%%%%%%%%%%%%%%%%%%%%%%%%%%%%%%%%%%%%%%%%%%%%%%%%
\section{Experiments}
\label{sec:Experiments}

\begin{figure}
\centering
\begin{tabular}{cccc}
\includegraphics[width=0.3\textwidth]{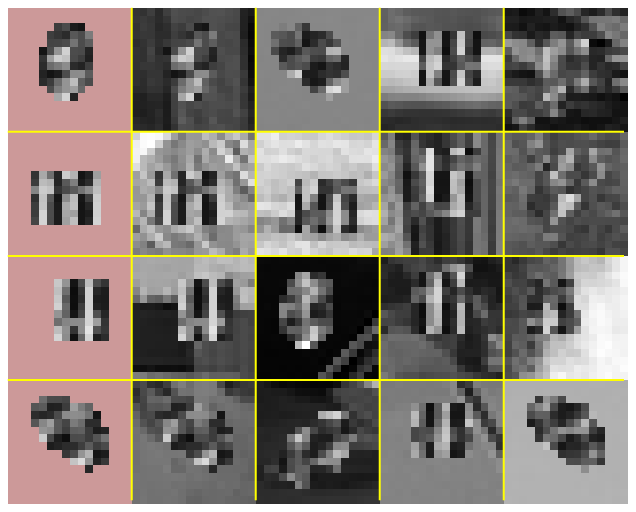}
&
\hspace{0.02\textwidth}
\includegraphics[width=0.19\textwidth]{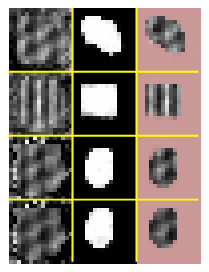}
&
\includegraphics[width=0.19\textwidth]{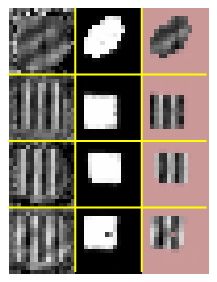}
\hspace{0.02\textwidth}
&
\includegraphics[width=0.246\textwidth]{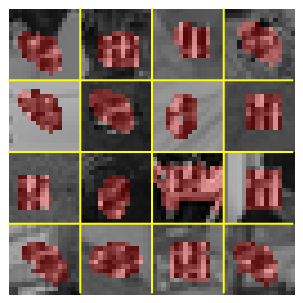}
\vspace{-0.2cm}
\\
{\tiny (a)} & \multicolumn{2}{c}{\tiny (b)} & {\tiny (c)}\\
\end{tabular}
\caption{
\textit{(a)} Training toy data: The two classes of ``objects'' are (1)
rectangles of 9 different sizes and (2) four kinds of round
shapes. Objects can appear at different locations in the image and are
filled with two different types of textures. The texture varies from
training image to training image but rectangles are always filled with
one type of texture and round objects with the other. The first
columns shows ``ground truth'', i.e. objects in isolation. The
remaining  columns show actual training data, i.e. objects embedded in
natural image backgrounds.
\textit{(b)} Samples from the model: In each block
the left column shows $\VF$ (appearance), the middle column shows $\m$
(shape) and the right column shows the joint sample where shape and
appearance have
been combined (red indicates the invisible part of the
sample). \textit{(c)} Test images with inferred masks $\m$
superimposed in semi-transparent red. The model largely identifies the
foreground objects correctly, but struggles sometimes, especially if
the background is poorly explained under the background model.}
\label{fig:ToySamples}
\end{figure}

\begin{figure}
\centering
\begin{tabular}{cc}
\multicolumn{2}{c}{\includegraphics[width=.85\textwidth, bb=192 411 411 434, clip=true]{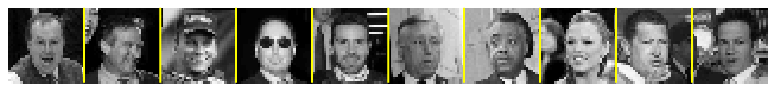}}
\vspace{-0.25cm}
\\
\multicolumn{2}{c}{{\tiny (a) }}
%\vspace{-0.1cm}
\\
\multicolumn{2}{c}{\includegraphics[width=.85\textwidth, bb= 200 381 407 464, clip=true]{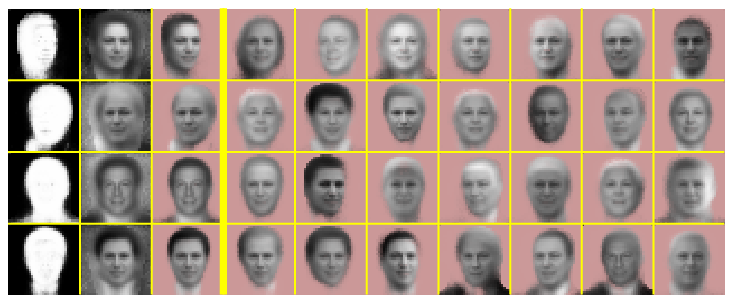}}
\vspace{-0.25cm}
\\\multicolumn{2}{c}{{\tiny (b) }}
%\vspace{-0.1cm}
\\
\multicolumn{2}{c}{\includegraphics[width=.85\textwidth, bb= 173 396 432 448, clip=true]{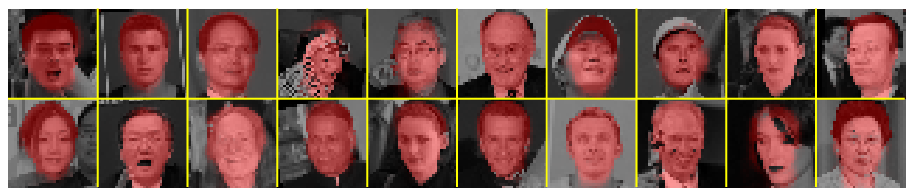}}
\vspace{-0.25cm}
\\
\multicolumn{2}{c}{{\tiny (c)}}
%\vspace{-0.1cm}
\\
\multicolumn{2}{c}{\includegraphics[width=.85\textwidth, bb=173 409 432 435, clip=true]{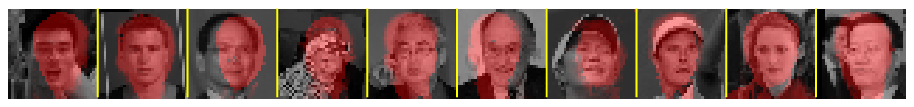}}
\vspace{-0.25cm}
\\ \multicolumn{2}{c}{{\tiny (d) }}
%\vspace{-0.1cm}
\\
\includegraphics[width=.43\textwidth, bb= 216 403 384 437, clip=true]{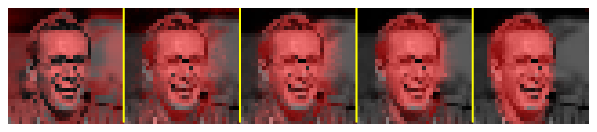}
&
\hspace{0.059\textwidth}
\includegraphics[width=.344\textwidth, bb= 221 403 380 444, clip=true]{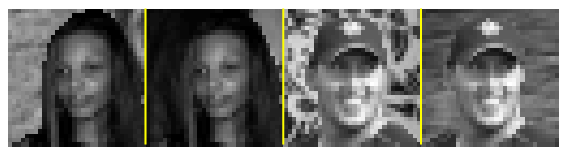}
\vspace{-0.25cm}
\\
{\tiny (e) } & {\tiny (f)}
\end{tabular}
\caption{
\textit{(a)} Examples of the training data. \textit{(b)}: Samples
from the learned model. For the first three columns the
format is similar to Fig.\
\ref{fig:ToySamples}b, they demonstrate how shape ($\m$, left) and
appearance ($\VF$, middle) combine to the joint sample (right). The remaining
columns show further samples from the model. For the joint samples the
red area is not part of the object. \textit{(c)}:
Inferred masks $\m$ (foreground-background segmentations) for a subset of
the test images. Masks are superimposed on the test images in red. In most cases the model has largely correctly identified the pixels
belonging to the face. Test images for which the model tends to make
mistakes typically show the head in extreme poses. Labeling of the neck
and the shoulders is somewhat inconsistent, which is expected
given that there is considerable variability in the training images
and that the model has not been trained to either include or exclude
such areas. The same applies if parts of a face are occluded,
e.g. by a hat. \textit{(d)} Test
images with random masks superimposed. If masks
are randomly assigned to test images the alignment of mask and image
is considerably worse.
Additional training images and segmentation results and a comparison
with results for a conventional conditional random field can be found
in the supplemental material \cite{suppl}. \textit{(e)}:
Convergence of the segmentation \textit{during
learning} (inferred mask $\m$ superimposed on training image before
joint training (left most) and after 10, 20, 100 and 1000 epochs of
joint training). At the beginning the segmentation is driven primarily by
the background model and thus very noisy. \textit{(f)} Two examples of the pairs of images used for the recognition task.}
\label{fig:facesResults}
\end{figure}

\textbf{Datasets \& evaluation}: we evaluate the model and the
learning scheme on two datasets: a toy dataset
and a more challenging, modified version of the ``Labeled faces in the
wild-A'' (LFW-A)-dataset~\cite{HuangTR2007,Wolf2009}. The toy data
consist of $16\times
16$ pixel image patches that contain two classes of foreground
``objects'' against backgrounds that are randomly cropped patches from
the VOC 2009 dataset. The two classes differ in appearance and shape
(i.e. shape and
appearance are \textit{dependent}) and objects can appear at various positions
in the patch (see  Fig.\ \ref{fig:ToySamples}).  For the
LFW-A dataset, the original images of size $250 \times 250$ were cropped to $210 \times
210$ pixels and down-scaled to $32 \times 32$ pixels. We used the
first 13000 images in the dataset for training and the remaining 233
images for test purposes. Examples are shown in Fig.\ \ref{fig:facesResults}a.
We evaluate the quality of learned models for both datasets by
sampling\footnote{
Markov chains are initialized with random noise; we use the conditional
means of the visible units given the  hidden units in the final step for
visualization. }
and by performing FG-BG segmentation on
unseen test images\footnote{We provide a comparison with the
  performance of a simple conditional random field (CRF) in the
supplemental material~\cite{suppl}.}.
We also investigate whether the FG-BG model's
ability to model the object of interest
independently of the background provides additional robustness of the
latent representation to background clutter in a recognition task.

\textbf{Models \& learning}: for our weakly supervised learning scheme, we first learned models of
the background by training Beta RBMs on large sets of natural image
patches using SML.
We next trained the foreground RBMs in the context of the full model
(eq.\ \ref{eq:FullModel} as described in section \ref{sec:Model}). For
each training data point, we store the state of the latent variables
$\VF$, $\VB$, and $\m$ from one epoch to the next. We update them by
one step of Gibbs sampling before using the state of the latent
variables for the current batch and the particles in the
persistent chains for the foreground model to compute the gradient
step for the parameters of the foreground RBM in the usual manner
\cite{TielemanICML2008}. Fig.\ \ref{fig:facesResults}e
illustrates how the mask $\m$ converges for a particular
\textit{training} image \textit{during learning} over 1000 epochs.
For the model to converge well on the face dataset we further
initialized the weights for the foreground appearance by training a
Beta RBM directly on the full face images (i.e. including background;
this pre-training does not teach the model to distinguish between
foreground and background in the training images).
For highly structured images the background
models were sometimes not sufficiently powerful so that part of the
background was assigned to the foreground even after consolidation
of the foreground model. This does not completely prohibit learning of
the foreground model but leads to a noisy final model.
We addressed this issue by introducing an outlier component
into the background model\footnote{
Using an ``outlier-indicator'' $o_i \in \{ 0, 1\}$ ($P(o_i = 1) =
p$) the constraints in eq.\ (\ref{eq:FullModel})
are replaced by  $\left(\prod_i \delta\left[\vF_i = x_i\right]^{m_i}
\delta\left[\vB_i =
x_i\right]^{(1-m_i)(1-o_i)}\left[U(x_i)\right]^{(1-m_i)o_i} \right)$.
For the toy
dataset we initially trained only with the basic model and introduced
the outlier component only for fine-tuning once the model had largely
converged. For the faces we trained with the outlier component from
the beginning.}, i.e. each background pixel was modeled
either by the background RBM or by a uniform distribution ($p =
0.3$),
which can be incorporated into the Gibbs sampling scheme by modifying
equations (\ref{eq:conditional1}-\ref{eq:conditional3}). Additional
details can be found in the supplemental material available on the first
author's homepage~\cite{suppl}.

%%%%%%%%%%%%%%%%%%%%%%%%%%%%%%%%%%%%%%%%%%%%%%%%%%%%%%%%%%%%%%%%%%%%%%
\section{Results}
\label{sec:Results}
%\begin{comment}
\textbf{Toy Data}:
For the toy data the model successfully learned about the shapes and
appearances of the foreground objects: after learning, samples closely
matched the foreground objects in the training set
(cf. Fig. \ref{fig:ToySamples}a) and inference on test patches led to
largely correct segmentations of these patches into foreground and
background: it labeled 96\% of the pixels in our set of 5000 test
images correctly
\footnote{
Chance is 50\%; a CRF using a histogram of the background and
contrast dependent smoothness term achieves 79\%; see supplemental
material~\cite{suppl} for further details.
}
(see Fig.\ \ref{fig:ToySamples}c for examples).
To investigate to what extent the ability to ignore the
background may help in recognition tasks we trained a simple RBM on
the same data (same total number of hidden units as for  the FG-BG
model) and performed inference for a set of test patches in both
models\footnote{Inference in the simple RBM involves
computing the activation of the hidden units given the test image; in
the FG-BG model it involves inferring $\m$, $\VF$, and $\VB$.
}. We trained a simple logistic classifier (with L2-regularization) on
the inferred activation of the hidden units (only $\HF$ for the FG-BG
model) in order to classify patches whether they contain rectangles or
round shapes. This an easy classification task
if enough training data is available (e.g. 100 training patches per
class), but as the number of training data points is reduced the
classification performance drops strongly for the simple RBM but to a
far lesser extent for the FG-BG model (classification performance 66\% vs.
88\% at 10 data points per class), suggesting that being able to
ignore the background can help to improve recognition performance.

\textbf{Faces}:
The model also learned a good representation of faces.
Fig.\ \ref{fig:facesResults}b shows samples from the
trained model. Although the samples do not exhibit as much detail as
the faces in the training data (this is to be expected given the relatively
small number of hidden units used)
they exhibit important features
of the training data, for instance, there are male and female faces,
and the model
has learned about different head positions and hair styles.
Figure \ref{fig:facesResults}c shows segmentation results for a subset
of the test images.
In most cases the model has largely correctly identified the pixels
belonging to the face. Test images for which the model tends to make
mistakes typically show heads in extreme poses.
Fig.\ \ref{fig:facesResults}d demonstrates that the model does not simply
choose the same region in all images: randomly re-assigning the
inferred masks to test images leads to considerably worse results.
To investigate the effect of the mask in a very simple recognition task we
manually segmented a small subset of the images in the dataset ($N =
65$). For each segmented person we created two test images by pasting
the person into two different natural image background patches, thus
obtaining two sets of images containing the same 65 different faces but
against different backgrounds (two example pairs are
shown in Fig.\ \ref{fig:facesResults}f). We inferred the
segmentation and subsequently the hidden activation of the latent
units of the foreground model. For each image from the first set we
determined the most similar image in the second set (in terms of
the RMS difference between the hidden unit activations). For 52 (80\%)
of the images in the first set the corresponding image with the same face
(different background) in the second set was the closest match. This
compares to 15 out of 65 (23\%) for a simple Beta RBM
(same number of hidden units as the foreground model) trained on the
same dataset, suggesting that here, too, ``ignoring'' the background
can lead to a representation in the hidden units that is less
affected by the background than for a normal RBM.

%%%%%%%%%%%%%%%%%%%%%%%%%%%%%%%%%%%%%%%%%%%%%%%%%%%%%%%%%%%%%%%%%%%%%%
%%%%%%%%%%%%%%%%%%%%%%%%%%%%%%%%%%%%%%%%%%%%%%%%%%%%%%%%%%%%%%%%%%%%%%
\section{Discussion}
\label{sec:Discussion}
We have demonstrated how RBMs and layered representations can be
combined to obtain a model that is able to represent the joint shape
and appearance of foreground objects independently of the background
and we have shown how to learn the  model of the foreground directly
from cluttered training images using only very weak supervision. The
architecture is very flexible: it can be applied to images of different
types (e.g. binary), the background model can be re-used for different
foreground models, and it is possible to replace the background model
independently of the foreground (e.g. if the statistics of the
background change). Also, DBNs or DBMs could be used instead of RBMs. One
interesting extension would be to include a third layer that is
\textit{in front} of the object layer. This would allow modeling
occlusion of the foreground object (such occlusions, although  rare in
the face dataset, may explain part of the uncertainty in the learned
shape model). Using \textit{semi}-supervised schemes (e.g. with a
few pre-segmented images) to learn more challenging object categories, is
another exciting direction.

\subsubsection*{Acknowledgments.}
NH is supported by a EPSRC/MRC scholarship from the Neuroinformatics
DTC at the University of Edinburgh.
NLR is supported by a grant from the European Research
Council (SIERRA-ERC-239993).

\end{document}